\documentclass[11pt,a4paper]{article}

\usepackage{authblk}

\usepackage[hyperref]{acl2021}
\usepackage{times}
\usepackage[shortlabels]{enumitem}
\usepackage{latexsym}

\usepackage{microtype}

\usepackage{comment}
\usepackage{mdframed}
\usepackage{multicol}
\usepackage{multirow}
\usepackage{booktabs}
\usepackage{graphicx}
\usepackage{tikz}
\usepackage{subfig}
\usepackage{xcolor,colortbl}

\newenvironment{itemize*}%
  {\begin{itemize}%
    \setlength{\itemsep}{0.9pt}%
    \setlength{\parskip}{0.9pt}%
    \setlength{\topsep}{0.9pt}}%
  {\end{itemize}}

\newenvironment{enumerate*}%
  {\begin{enumerate}%
    \setlength{\itemsep}{0.9pt}%
    \setlength{\parskip}{0.9pt}%
    \setlength{\topsep}{0.9pt}}%
  {\end{enumerate}}

\aclfinalcopy 
\def\aclpaperid{2342} 

\title{On the Interaction of Belief Bias and Explanations}

\author[1]{\textbf{Ana Valeria Gonz{\'a}lez}}
\author[2]{\textbf{Anna Rogers}}
\author[1]{\textbf{Anders Søgaard}}
\affil[ ]{University of Copenhagen}
\affil[1]{Department of Computer Science}
\affil[2]{Copenhagen Centre for Social Data Science}
\affil[ ]{\texttt{\{ana,soegaard\}@di.ku.dk}}
\affil[ ]{\texttt{arogers@sodas.ku.dk}}

\date{}

\begin{document}
\maketitle
\begin{abstract}
A myriad of explainability methods have been proposed in recent years, but there is little consensus on how to evaluate them. While automatic metrics allow for quick benchmarking, it isn't clear how such metrics reflect human interaction with explanations. Human evaluation is of paramount importance, 
but previous protocols fail to account for \textit{belief biases} affecting human performance, which may lead to misleading conclusions. We provide an overview of belief bias, its role in human evaluation, and ideas for NLP practitioners on how to account for it. 
For two experimental paradigms, we present a case study of gradient-based explainability introducing simple ways to account for humans' prior beliefs: models of varying quality and adversarial examples. We show that \textit{conclusions about the highest performing methods change when introducing such controls}, pointing to the importance of accounting for belief bias in evaluation. 

\end{abstract}

\section{Introduction}

Machine learning has become an integral part of our lives; from everyday use (e.g., search, translation, recommendations) to high-stake applications in healthcare, law, or transportation. However, its impact is controversial: neural models have been shown to make confident predictions relying on artifacts \cite{mccoy-acl19,wallace2019universal} and have shown to encode and amplify negative social biases \cite{manzini2019black, caliskan2017semantics,may-etal-2019-measuring, tan2019assessing, gonzalez2020type, rudinger2018gender}.

\begin{figure}[h]
\centering
\includegraphics[width=.4\textwidth]{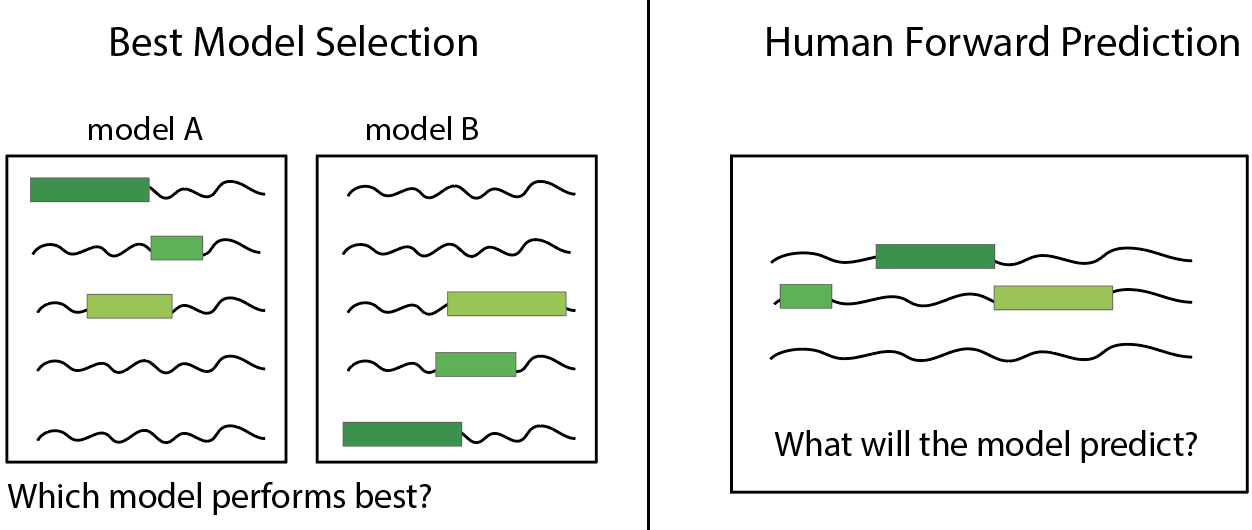}

\caption{\label{fig:protocol} Evaluation protocols considered in this work} 
\end{figure}





\textit{Explainability} aims to make model decisions transparent and predictable to humans; it serves as a tool for model diagnosis, detecting failure modes and biases, and more generally, to increase trust by providing transparency \cite{amershi-chi19}. 
While automatic metrics have been proposed to evaluate various properties of explanations such as faithfulness, consistency and agreement with human explanations \cite{atanasova2020diagnostic,robnik2018perturbation, deyoung-etal-2020-eraser}, these metrics do not inform us about human interaction with explanations.

\citet{doshi2017towards} suggested \textit{human forward prediction}, a simulation task in which humans are given an input and an explanation, and their task is to predict the expected model output, regardless of the gold answer. Recent studies include \citet{nguyen2018comparing,lage2019evaluation, hase-bansal-2020-evaluating, poursabzi2018manipulating}. Such protocols are widely used and can provide valuable insight into human understanding of explanations. However, prior work has not accounted for how humans' prior beliefs (\textit{belief biases}) interact with the evaluation; simulating model decisions becomes an easier task when the model being evaluated makes predictions which align with human expectations. We argue that not considering belief bias in such protocols \textit{may lead to misleading conclusions about which explainability methods perform best}.

Other protocols have evaluated participant’s ability to select the best model based on explanations offered by different interpretability methods (e.g. decide which model would generalize ‘in the wild’) \cite{ribeiro2016should}. 
However, comparisons have been made between a model which is clearly in line with human beliefs, and another which exploits spurious correlations diverging from human expectations. When differences are less obvious, humans may not be able to leverage their belief biases, and conclusions may change. 


This paper, which includes evaluations for both of the previously mentioned tasks, closes an important gap: to the best of our knowledge, no prior work in NLP addresses the interaction of belief bias with current human evaluations of explainability.

\paragraph{Contributions.}  We provide an overview of belief bias meant to highlight its role in human evaluation and provide some preliminary ideas for NLP practitioners on how to handle such cases. 
Using \textit{human forward prediction} and \textit{best model selection} (\autoref{fig:protocol}), we present a case-study where we compare two gradient-based explainability methods in the context of reading comprehension (RC), introducing conditions to take into account belief bias. We find that both explainability methods are helpful to participants in the standard settings (in line with most previous work), but the \textit{conclusions about the best performing models change when incorporating additional control conditions}, reinforcing the importance of accounting for such biases.

\section{Belief Bias}
\label{sec:beliefbias}
\textbf{Belief bias} is a type of cognitive bias, defined in psychology as \textit{the systematic (non-logical) tendency to evaluate a statement on the basis of prior belief rather than its logical strength} \cite{evans1983conflict,Klauer:ea:00, barston1986investigation}. Cognitive biases are not necessarily bad; they help us filter and process a great deal of information \cite {bierema2020quantifying}, and have been widely studied in real human-decision making \cite{tversky1974judgment,kahneman2003perspective,furnham2011literature}. However, in evaluations involving human participants, such biases may alter results and affect conclusions \cite{anderson2014belief,wall2017warning}.

Classic psychology studies of belief bias have assessed how prior beliefs affect syllogistic reasoning \cite{newstead1992source,Klauer:ea:00,evans1983conflict,markovits1989belief,evans54sj}. Consider the following example by \citet{anderson2014belief}: 

\begin{mdframed}
\begin{enumerate}[(a)]
    \setlength{\itemsep}{0.9pt}%
    \setlength{\parskip}{0.9pt}%
    \setlength{\topsep}{0.9pt}%
\footnotesize
 \item \textit{If all birds are animals, and if no animals can fly, then no birds can fly.} 
 \item \textit{If all cats are animals, and if no animals can fly, then no cats can fly.}     
\end{enumerate}
\end{mdframed}

In syllogistic reasoning, the task for humans is to assess the \textit{logical} validity of such arguments while ignoring believability. While both arguments are logically valid, most work converges on the finding that humans will rate argument (a) as invalid more often than (b), biased by the fact that the premise in (a) is less believable.


In psychology, belief bias has been tied to the \textit{dual-processing} theory, which assumes that reasoning is performed by two competing cognitive systems: (1) \textit{system 1} which takes care of fast, heuristic processes and (2) \textit{system 2} which handles slower, more analytical processes \cite{evans2003two,trippas2018parallel,evans2005rapid,croskerry2009universal}. Generally, humans tend to have a cognitive preference for relying on fast, intuitive \textit{system 1} processes, rather than engaging in the slow and more analytical \textit{system 2} processes.
 Belief bias is attributed to system 1 \cite{evans2005rapid,evans2008dual,evans2009two,stanovich2008relative} due to several factors, reviewed in detail by \citet{evans2003two,caravona2019get}. 

For the purposes of NLP studies relying on crowd workers, one relevant finding is that \textbf{time pressures exacerbate reliance on previous beliefs} \cite{evans2005rapid}. Since crowd workers generally are incentivized to work as quickly as possible to maximize their hourly pay, reliance on belief bias is to be expected.

Another relevant finding for NLP is that threatening or negatively charged arguments (e.g. content violating political correctness and social norms) leads to greater engagement of system 2, whereas \textbf{neutral content leads to increased reliance on belief bias} 
\cite{goel2011negative,klaczynski1997goal}. Since NLP studies tend to be performed on neutral content such as passages from Wikipedia -- content which may not sufficiently engage participants' system 2 processes -- belief bias is more likely to play a role in human performance.



This study aims to highlight the phenomenon of belief bias to encourage NLP practitioners to assess the role it plays in their evaluations, and introduce mechanisms to account for belief bias effects. We illustrate how belief bias effects can significantly affect the results of human evaluation of explainability for two paradigms: \textit{human forward prediction} and \textit{best model selection}.

\section{Related Work}


\paragraph{Human forward prediction.}

Human forward prediction experiments have been recently presented in the context of synthetic data \cite{poursabzi2018manipulating,lage2019evaluation,slack2019assessing} to evaluate explainability methods for \textit{their ability to make model decisions predictable to humans}. In this paradigm, humans are presented with explanations and tasked with predicting the model's decision regardless of the ground truth \cite{doshi2017towards}.\footnote{Using synthetic data from fictitious domains effectively controls for belief bias \cite{lage2019evaluation,slack2019assessing}. \newcite{slack2019assessing}, for example, evaluate explanations in the domain of recommending recipes and medicines to aliens.} 

In NLP, \citet{nguyen2018comparing} introduced human forward prediction for LIME explanations \cite{ribeiro2016model} of sentiment analysis of product reviews and correlated the results with automatic evaluations. Unlike with synthetic data, participants have prior beliefs on what the \textit{true} outcome is. Since participants in \newcite{nguyen2018comparing} had no training phase to learn how explanations correlate with predictions and the model being evaluated sufficiently matched human behavior, humans likely relied {\em exclusively} on their prior knowledge and beliefs to complete the task at hand.   

\citet{hase-bansal-2020-evaluating} 
improved on this protocol by adding a training phase. This is something we also do in our experiments (\autoref{sec:hfp}), 
but it is unlikely to solve the belief bias problem because even after training, humans will naturally opt for fast, heuristic mechanisms (e.g. belief bias) in order to simplify tasks \cite{wang2019designing}; this is particularly true if the model is high performing (i.e. likely aligns with human beliefs).

The protocol by \citet{hase-bansal-2020-evaluating} had another key feature: they leave out the explanations for the test data points. This would seem like an advantage for evaluating explainability methods in the context of reading comprehension where explanations can, in theory, simply highlight the answer span, making it easy to guess the model output from the explanations. However, it is easy to control for the amount of explanation provided by the explanation methods we compare; in our experiments below, we highlight the top 10 tokens with highest attribution scores. This key feature in their protocol is problematic for two reasons:
\begin{itemize*}
\item It makes the human learning problem much harder, and we argue it is infeasible to expose participants to enough examples to make human forward prediction learnable (unless the task is made very easy on purpose; again by only evaluating high performing models). If it is not learnable, participants fall back on belief bias.
\item It introduces a systematic bias between the training and test scenarios.
\end{itemize*}

The protocol in \citet{hase-bansal-2020-evaluating} also does not randomize the order in which participants are exposed to problems with or without explanations. 

We improve on the above protocol by introducing a condition which can help account for belief bias effects: evaluating explainability methods on low-quality models, the predictions of which substantially differ from human beliefs.  
This means that in order to succeed in the task, humans cannot simply rely on their previous beliefs, therefore, helping us assess the ability of explanations in helping humans to \textit{realign} their expectations of model behavior. The predictions of reading comprehension models can also be made different from human answers by introducing distractor sentences that fool machine reading models, but not humans 
\cite{jia2017adversarial}. If in human forward prediction, participants predict the true answer rather than spans in the distractor sentences, this suggests 
participants may be relying on their belief biases.

\paragraph{Best model selection.}
\citet{ribeiro2016model} presented an evaluation of explainability methods for text classification, where explanations for decisions of two different models on the same instance are presented side by side, and humans decide which model is likely to generalize better. With some exceptions \cite{lertvittayakumjorn2019human}, there has not been much follow up work on this task, but this scenario is important: it mimicks the decisions about what model is \textit{safer for deployment}. \citet{ribeiro2016model} and  \citet{lertvittayakumjorn2019human} both make a single comparison between a model which clearly diverges from human intuition, and a model that generalizes and \textit{aligns with humans' beliefs}. Accounting for the extent to which belief biases are leveraged (e.g. by introducing additional model comparisons where differences are not so obvious or where models are of low quality) is important in such paradigms, and can allow us to better evaluate where explanation methods may fail.

In the following sections, we show that introducing conditions which take into account belief biases can have an effect on the conclusions for both \textit{human forward prediction} and \textit{best model selection}. We emphasize that many other potential strategies can be introduced and this is largely dependent on the goals of the evaluation protocol; we merely provide one example case with the following strategies:

\begin{itemize*}
\item[(1)] Introducing low quality models which considerably diverge from humans' prior beliefs (\textit{human forward prediction})

\item[(2)] Introducing evaluation problems with distractor sentences (\textit{human forward prediction})

\item[(3)] Introducing model comparisons where relying on belief bias is not enough to obtain high performance (\textit{best model selection})
\end{itemize*}

\section{Experimental Setup}
This section introduces the general setup of the experiments, with details specific to each experimental paradigm described in \autoref{sec:hfp} and \autoref{sec:bms}.

\subsection{Models}

We evaluate explanations produced by three BERT-based \cite{devlin-etal-2019-bert} models:

\begin{itemize*}
    \item[(a)] a high performing model ({\sc \textbf{High}}): BERT-base, fine-tuned on SQuAD~2.0. \textit{This model is more aligned with human beliefs.}
    
    \item[(b)] a medium performing model ({\sc \textbf{Medium}}): tinyBERT, a 6-layer distilled version of BERT \cite{jiao2019tinybert}, fine-tuned on SQuAD~2.0. 
    It performs about 20 $F_1$ points below {\sc High}. \textit{This model somewhat aligns with human intuition, but performs significantly lower.}
    
    \item[(c)] a low performing model ({\sc \textbf{Low}}): BERT-base, fine-tuned to always choose the first occurrence of the last word of the question. This system mimicks a rule-based system\footnote{This model achieves about 0.90 F1 for this task, but in the results we show its performance on the actual RC task}; however, we evaluate gradient-based methods requiring a neural model. \textit{This model diverges significantly from human beliefs.}
    
\end{itemize*}

\subsection{Data}

We use SQuAD~2.0 \cite{rajpurkar2018know}, a RC dataset consisting of 150k factoid question-answer pairs, with texts coming from Wikipedia articles. We opt for this data as it contains short passages that can be read by humans in a short time. In the human forward prediction experiments, we refer to experiments using this data as {\sc orig}. As described in \autoref{sec:beliefbias}, Wikipedia texts could by themselves induce people to rely on their belief bias, but this particular dataset allows us to also introduce controls for the bias: the adversarial version of the data \cite{jia2017adversarial}, has been shown to distract models but not humans. This means that in order to perform the task with success, humans need disregard their belief biases, and in some cases align with distractor sentences. We refer to this data in our simulation experiments as {\sc Adv}.



\subsection{Explainability Methods}
We focus on gradient-based approaches, as they require no modifications to the original network, and are considerably faster than perturbation-based methods. 
We compare two explainability methods:

\paragraph{Gradients.}
Computing the gradient of the prediction output with regard to the features of the input is a common way to interpret deep neural networks \cite{simonyan2013deep} and capture relevant information regarding the underlying model.  

\paragraph{Integrated gradients.}
Integrated gradients approach (IG) \cite{sundararajan2017axiomatic} attributes an importance score to each input feature by approximating the integral of gradients of the model’s output with respect to the inputs along the path, from the references to the inputs. 
IG was introduced to address the sensitivity issues which are present in vanilla gradients and implementation invariance. 

\section{Experiment 1: Human Forward Prediction}
\label{sec:hfp}
Human forward prediction for evaluating explainability was proposed by \citet{doshi2017towards}. They argue that if a human is able to simulate the model's behavior, they understand \textit{why} the model predicts in that manner. For the reasons previously outlined, we suspect that belief biases may be affecting performance and the conclusions once can draw from this task. We investigate this by asking the following: \textit{Can humans predict model decisions, if model behavior considerably diverges from their own beliefs?} 

     


\paragraph{Stimuli presentation.} 
We include: (i) {\sc High}, which is finetuned to solve SQuAD~2.0 and (ii) {\sc Low}, which is finetuned to select the first appearance in the context of the last word in the question. We evaluate each of the two models twice: with or without adversarial data. We contrast using vanilla gradients and IG with a baseline condition, in which no explanations are shown ({\sc Baseline}).

We highlight the top-10 tokens\footnote{Explanations should be {\em selective} \cite{mittelstadt2019explaining}} with the highest attribution scores wrt. the start and end positions of the predicted span, and zero out the rest.\footnote{\citet{ribeiro2016should} use the top 6 attributes; we opt for 10 given that our texts are slightly longer.} The two sets of tokens often overlap. 

\begin{figure}[t!]
\centering
\includegraphics[width=.42\textwidth]{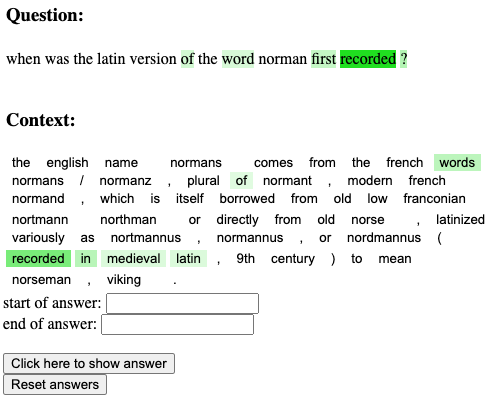}

\caption{\label{fig:e2-ui} Interface for Experiment 1 for {\sc low} condition. To select model predictions, participants clicked on tokens to select the start and end of the span. Then they would see the actual model prediction.}
\end{figure}

Participants were provided with a question and a passage (with or without explanations) and were told to pick the \textit{shortest} span of text which matched the model prediction. They saw the actual model answers before the next example (done for both baseline and explanation conditions), which was an important part of training to infer model behavior. Before the model prediction was shown, their answers were locked to prevent any further changes. 
An example of our interface can be found in \autoref{fig:e2-ui} and the instructions are shown in \autoref{e2-app}.

We ran these experiments on Amazon Mechanical Turk, recruiting participants with approval ratings greater than 95\%\footnote{Previous research has shown that proper filtering and selection of participants on Mechanical Turk, can be enough to ensure high quality data \cite{peer2014reputation}.} and ensuring different groups of participants per condition by specifying that participation is only allowed once, otherwise risking rejection\footnote{We also remove such (few) repetitions at analysis}. We paid participants \$5.25 for about 20 minutes of work (to ensure at least a \$15 hourly pay) and obtained at least three annotations per example.  The data included 120 unique questions divided into small fixed batches (the same questions across conditions). 
About 75\% of questions are accurate in the {\sc High} model, and around 15\% are accurate for the {\sc Low} model. 
In total, we obtained 4,300 data points across 123 participants (35 data points per participant).

\paragraph{Results.} As humans often did not select the exact span that was provided as ground truth, we \textit{manually} labeled the spans as correct or incorrect. We also inspected the impact of training in human forward prediction, e.g., the learning effect of multiple exposures on annotator accuracy. Both with vanilla gradients and integrated gradients, we observe an increase in the participants' accuracy at around 15 examples. In contrast, in our baseline condition, performance either stays constant or drops slightly. To reduce the noise introduced due to the training period, we remove the first 15 examples of each participant. The results without this preprocessing (\autoref{e2-app}) suggest that \textbf{the effect of training differed across explainability methods}, as will be discussed later in the section. 

Using the average human accuracies per example, we run a one-way ANOVA to test for significant differences across the groups. As we obtained statistically significant results, we then ran the Tukey honest significant difference (HSD) test \cite{tukey1949comparing}, comparing the means of every condition to the means of every other condition. The results are presented in \autoref{tab:main_results}. 

As expected, in the absence of explanations ({\sc baseline}), \textbf{humans rely on belief bias and predict the gold standard answer more often than the model prediction} 
($y$ in \autoref{tab:main_results}). Even with training (seeing the true model prediction), humans fail to catch onto the simple rule used by the {\sc Low} model, when no explanations are presented.

\begin{table}[t]
\footnotesize
\begin{center}
\begin{tabular}{l| r |rrr}
\toprule

&{\sc Model}&\multicolumn{3}{c}{\sc Human}\\
\midrule 
{\sc condition}& {\sc F1}  & {\sc $\hat y$ } & {\sc $y$} &{\sc sec}  \\
\midrule 

& & \multicolumn{3}{c}{\sc \textbf{Baseline}} \\
\midrule 
 {\sc Low-orig}& 0.17
 & 0.16 & 0.48 &33.9 \\
 {\sc Low-Adv}& 0.15
 & 0.12 & 0.34&63.3\\
 \midrule
 {\sc High-orig} & 0.79
 & 0.45& 0.46 &34.6\\
 {\sc High-Adv}& 0.66%
  & 0.38& 0.48 & 36.1\\
 \midrule 
& & \multicolumn{3}{c}{\sc \textbf{Integrated (IG)}} \\
\midrule 
 {\sc Low-orig}& \cellcolor{gray}& $^*$0.58 & $^*$0.22 &$^*$16.8 \\
{\sc Low-Adv}& \cellcolor{gray}& $^*$0.63 &$^*$0.18 &$^*$22.3 \\ 
\midrule 

{\sc High-orig} &\cellcolor{gray} &  $^*$\textbf{0.84 }& $^*$0.88 &36.1  \\
{\sc High-Adv}& \cellcolor{gray}&  $^*$\textbf{0.52}&$^*$ 0.35 & $^*$18.9 \\
\midrule 
& & \multicolumn{3}{c}{\sc \textbf{Gradients}} \\
\midrule 
 {\sc Low-orig}& \cellcolor{gray}&  $^*$\textbf{0.69} & $^*$0.06 &32.6\\
{\sc Low-Adv}& \cellcolor{gray}& $^*$\textbf{0.72} &$^*$0.15 &$^*$25.6\\ 

\midrule 

{\sc High-orig} &\cellcolor{gray} & $^*$0.79 & $^*$0.81 &47.4  \\
{\sc High-Adv}& \cellcolor{gray}&  0.49& $^*$0.60 &48.4 \\
\bottomrule
\end{tabular}
\caption{Human forward prediction results ({\sc Human}($\hat y$)) for {\sc Low} and {\sc High} models, compared to no explanations ({\sc Baseline}). Each experiment is run on vanilla SQuAD~2.0 data ({\sc Orig}) and adversarial SQuAD~2.0 data ({\sc Adv}). {\sc Human}($y$) is the dataset ground truth and an indicator of belief bias. Statistically significant results are indicated with an asterisk. Time is the average time per question. The best $\hat y$ results in each condition are bolded.}
\label{tab:main_results} 
\end{center}
\end{table}

Overall, explanations derived from both of the gradient-based approaches lead to statistically significant improvements over the baseline. This indicates that the \textbf{explanations allow humans to realign their expectations of the model behavior}, better than with no explanations. 

For {\sc HIGH-ORIG}, the standard setting explored in previous evaluations, both IG gradients and vanilla gradients perform well, with IG gradients performing better. Given these results and the theoretical advantages of IG over vanilla gradients, one could arrive at the conclusion that IG are better for simulatability. 
However, \textbf{the differences between the two gradient-based methods are reversed in the conditions where humans cannot rely on their previous beliefs} ({\sc Low}). The gap between gradients and IG as large as 0.11, and being statistically significant. This finding is \textit{surprising} and points again to the importance of not drawing incorrect conclusions about the best performing method using the standard paradigm.  

Finally, in the {\sc high} conditions, model behavior decreases about 13\%~F1 score with the presence of \textbf{adversarial examples}, meaning that the model we used does get affected by adversarial inputs. We observe that human performance is considerably lower in {\sc high-adv} as opposed to {\sc high-orig}. \textbf{With vanilla gradients, performance is more aligned with the ground truth labels than with model behavior,} showing that in this condition humans are also relying on their prior beliefs. \textbf{With IG, where performance is less aligned with prior beliefs (ground truth), the end performance increases}, but it seems that this condition is considerably more difficult for humans.

\paragraph{Effect of training.}
In {\sc baseline}, training does not affect either the {\sc low} or {\sc high} conditions (see \autoref{tab:raw} in \autoref{e2-app} for the raw results). For the {\sc low} model, multiple factors can be taking place (possibly at the same time): (1) the task is too far from the humans' beliefs and there is no mechanism to help participants realign their expectations, (2) participants may not be incentivized to seriously engage and look for patterns, (3) participants opt for a mixed strategy, where for some questions they go with their prior beliefs and for others, choice is random (as seen in their performance in $y$). 

For {\sc high} conditions in {\sc baseline}, performance remains higher than {\sc low} but this is likely due to belief bias and not training, given that performance remains constant after removing the training data points. We hypothesize that for {\sc high}, instances where the model does not align to human intuition might be more detrimental than in explanation conditions. More specifically, if humans are aware that the model aligns with their beliefs after some examples but encounter instances where it doesn't (model is not 100\% accurate), they will likely develop an expectation that the model is bound to make some errors, without any indication of when. 

In addition, our raw results suggest IG required longer training. While this does not mean IG is a worse method than vanilla gradients, explanations derived from IG may have confused participants due to containing information which was irrelevant to them. It may be that experts (e.g. system engineers knowledgeable about neural networks) can take better advantage of such explanations; however, we leave this exploration of the interaction of human expertise with explanations as a direction for future work.

\section{Experiment 2: Best Model Selection}
\label{sec:bms}
This section presents the setup and results of our model selection experiments; a task where humans select the model that is more likely to succeed in the wild. We present the participants with the explanations from two models ({\sc High} vs {\sc Low} and {\sc High} vs {\sc Medium}), and ask them to decide which model is likely to perform better. As a follow-up, we also experimented with \textit{soliciting explanations about what leads the worse model to fail}. Intuitively, comparative evaluation difficulty depends on  how clear the difference is between the compared objects. 
Explanations should at least show the difference between a high-performing model  and a low-performing one, enabling human participants to predict which is  better (standard setting).  




\paragraph{Stimuli presentation.} 
We presented participants with saliency information from both models (a high performing model + one of the lower performing models), and their task was to determine which model performs best in the wild. We shuffled the order at random so that the best model would not remain in a fixed position. We obtain 120 samples (question-context pairs), and show the explanations next to each other as seen in \autoref{fig:protocol}. The participants are told that the highlighted attributes are the words the model found important in making its  decision. A screenshot of the UI is shown in \autoref{fig:e1} in \autoref{instructions1} and the instructions provided to the participants are also shown in \autoref{instructions1}. These experiments were also ran on Amazon Mechanical Turk with the same general procedures and pay.  The same subset of 120 examples is used in all conditions.  We obtained at least three annotations per example and ended with a total of 1440 data points across 48 participants (30 examples each).

\begin{figure*}[t]%

    \centering
    \includegraphics[width=.8\textwidth]{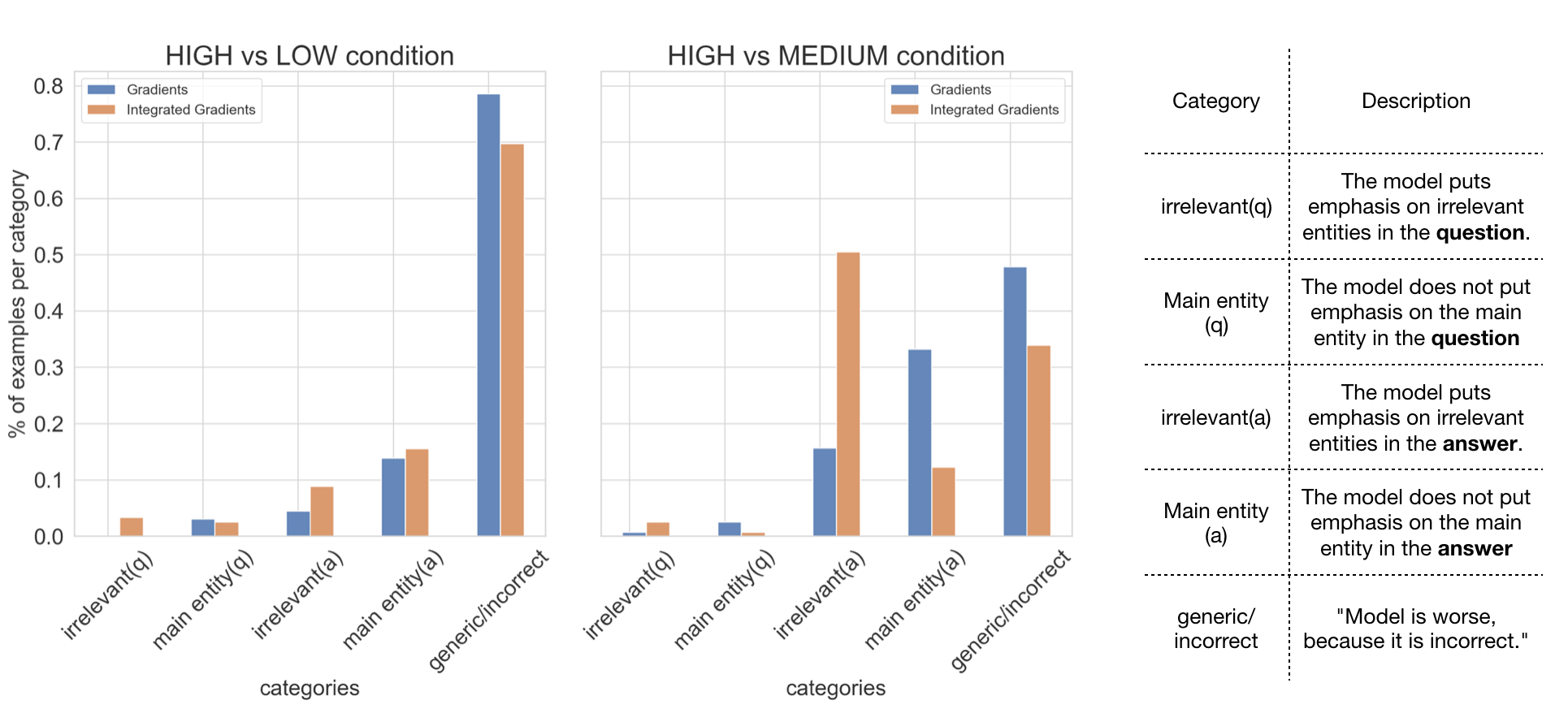} %

    \caption{Feedback categories and their distribution. We observed that the {\sc High} vs {\sc Medium} condition results are considerably different from the {\sc High} vs {\sc Low} condition, with more participants giving generic answers for vanilla gradients, and emphasizing the irrelevant terms highlighted in the IG condition.}
    \label{fig:distribution}%
\end{figure*}

\paragraph{Results.} For each example shown to annotators, we obtained the average accuracy scores and performed a standard T-test to compare the performance of the two methods. The results are shown in \autoref{tab:exp-1}. Using explanations from both methods, when shown the {\sc High} and {\sc Low} model, humans are clearly able to correctly select the better one.  With IG, humans achieve \textbf{0.95} accuracy on average, while with vanilla gradients they achieve \textbf{0.89}. The difference is {\em not} statistically significant. 
The fact that users are consistently able to discriminate between {\sc High} and {\sc Low} models is expected, and serves as a \textit{sanity check} that these explanations are meaningful for humans.

\begin{table}[h]
\begin{center}
\footnotesize
\begin{tabular}{l|cc}
\toprule

{\footnotesize Condition} & {\footnotesize Gradients} & {\footnotesize IG}\\

\midrule 

{\sc \footnotesize High vs low} & {\footnotesize 0.89} & {\footnotesize 0.95}\\
{\sc \footnotesize High vs medium*} & {\footnotesize 0.85} & {\footnotesize 0.52}\\
\bottomrule
\end{tabular}
\end{center}
\caption{\label{tab:exp-1} Both methods do well in ({\sc High vs Low}). In {\sc high vs \sc Medium}, performance drops dramatically for IG. * = statistical significant difference ($\rho < 0.001$)}
\end{table}

When the same experiment was repeated in the {\sc High} vs {\sc Medium} condition, we found clear and statistically significant differences between the two explainability methods. Using IG, participants reach only \textbf{0.52} accuracy, while with vanilla gradients their performance is \textbf{0.85}. This is \textit{surprising}, given that the difference in performance between the two models is still quite large (about 20\%~F1); the expectation is that both methods would capture this difference relatively well. It appears that when both models \textit{more or less} align with human beliefs, the task is much more difficult. To solve the task, humans now need to engage in more analytical thinking and cannot simply rely on belief biases to solve the task. We further investigate these differences through qualitative coding.

\paragraph{Qualitative analysis.} After each instance, we asked participants to describe how the worse model will fail. We do not provide detailed guidelines in order to not further bias the participants by introducing specific criteria. The instructions given to the participants are shown in \autoref{e1-app}. 

We collected 1440 responses, which were all inspected manually to uncover categories (codes). After multiple iterations, we tagged each response with one code (categories are mutually exclusive, no response can be placed in two). A description of the categories and their distribution are shown in \autoref{fig:distribution}, and examples of feedback per category are provided in the \autoref{e1-app}.

In the {\sc High} vs {\sc Low} condition, feedback for both methods was generic (about 70-80\% of the time), e.g., \textit{model B is likely incorrect so it is worse}. This was expected: this task should be easy when model differences are large and humans can rely on their \textit{system 1} processes to get through the task without thinking deeply about the explanations. 

In the {\sc High} vs {\sc Medium} condition, the distribution of the feedback categories is very different. For IG, 50\% of the time participants \textit{felt} the highlighted tokens where irrelevant.  This is not the case for gradients, where only about 15\% of responses fell in that category. Additionally, for vanilla gradients, 50\% of feedback is generic, signaling that in this condition, it may have been an easy task as well; explanations are making model behavior clear enough. It remains an open question whether IG explanations may in fact be more faithful to the model reasoning. In that case, \textit{expert users} (e.g. a system engineer debugging a system) may not find IG attributions irrelevant and would be able take better advantage of the information provided. For this reason, other kinds of human participants may show different results. Nevertheless, as evaluating on non-experts (crowdsourced workers for example) is common, this preliminary result is important: it shows that \textbf{conclusions can shift dramatically when introducing additional model comparisons} which reduce the participants' ability to rely on prior knowledge. 


\section{Discussion: Mitigating Belief Bias}

This study introduced \textit{additional conditions} in which the human participants could not rely on their belief biases to facilitate the task at hand. We presented a case study on evaluating reading comprehension models in model selection and human forward prediction paradigms, and we showed that this simple addition led to different conclusions in the evaluation and a better understanding of how humans interacted with explanations. Other tasks and paradigms might call for different setups, but generally including conditions with models of varying quality would be helpful both for the purposes of bias control, and for simulation of real-life use of explainability techniques to support decisions about which model is safer to deploy.

To conclude, we will briefly mention other directions for mitigating belief biases that can also be explored in future work and which should be kept in mind when developing evaluation protocols for explainability.



\paragraph{Reducing ambiguity.}
Ambiguity of task instructions leads humans to align interpretations to their own prior beliefs \cite{heath1991preference}; this may lead to misinterpretation and results which do not reflect the intended interaction with explanations. Ambiguity may also be present in other parts of the evaluation setup. For example, \citet{lamm2020qed} evaluate the effectiveness of explanations in helping humans detect model errors for open-domain QA, but the data they use contains questions where multiple answers can be true. Users may deem an answer to be correct or incorrect based on their understanding of the question, which makes the effect of explanations blurry. Removing ambiguous instances from the data can be a way of reducing such confounds.

\paragraph{Removing time constraints.}
Time constraints exacerbate reliance of system 1 processes, which leads to humans relying on belief biases. In crowdsourced evaluations, it is common practice to to provide workers with enough time to perform tasks, but workers may have intrinsic motivations for performing tasks quickly. A major challenge for evaluation research with crowd workers is creating better incentives for engaging in system 2 processes, e.g. pay schemes which encourage workers to be more analytical and accurate \cite{bansal2019updates}.

\paragraph{Include fictitious domains.} 
Using data from domains from which subjects have no prior beliefs e.g. fictitious domains, may be an efficient way of controlling for belief bias in some tasks\footnote{Again, we emphasize that some strategies are task dependent; fictitious domains may not be relevant in some tasks.}. This strategy has been used outside of NLP \cite{poursabzi2018manipulating,lage2019evaluation,slack2019assessing}, where subjects are asked to imagine alternative worlds such as scenarios involving aliens. In QA for example, one could introduce context-question pairs that describe facts about fictitious scenarios that sufficiently differ from human reality. 

\section{Conclusion}
The main contribution of this paper is bringing the discussion of belief bias from psychology into the context of evaluating explainability methods in NLP. Belief bias is a phenomenon which plays a role in human decision making and which interacts with previous evaluations in a way which may affect the conclusions we draw from these paradigms. 
We provide an overview of belief bias, making a connection between findings in psychology and the field of NLP, and present a case study of evaluating explanations for BERT-based reading comprehension models. We show that introducing models of various quality and adversarial examples can help to account for belief bias, and that introducing such conditions affects the conclusions about which explainability method works better. Finally, we provide additional insights and ideas for how to account for belief bias effects in human evaluation.

\section{Broader Impact Statement}

The work presented here makes strides towards a better understanding about the interaction of humans with explanations of model decisions. We have highlighted a phenomenon studied in psychology with hope that this opens the door to more NLP research involving a wider and more interdisciplinary understanding of humans, and the effect of explainability. 

This study involved human participants recruited on Mechanical Turk platform. No personally identifiable data was collected from the participants, they were made aware that the data would only be used for research, and they were not exposed to any emotionally traumatizing or offensive stimuli. We ensured a minimum \$15 hourly wage.

\section*{Acknowledgements}
We thank the reviewers for their insightful feedback for this and previous versions of this paper. This work is partly funded by the Innovation Fund Denmark.

\bibliographystyle{acl_natbib}
\bibliography{anthology,acl2021}

\appendix
\label{sec:appendix}
\clearpage

\section{Experiment 1: Human Forward Prediction}
\label{e2-app}
Below we show the instructions provided to the participants, as well as an example of the saliency maps presented to participants for adversarial examples.

\paragraph{Instructions.}
Question-answering systems are a particular form of artificial intelligence. The task here is for you to learn to predict how the system answers questions. In other words, when in a bit, you are presented with questions, the task is not to provide the right answer, but to guess the answer the system provided. For each question, you will also see a context paragraph. The answer is a span of text in this paragraph. Instead of writing out the answer, you can simply mark the relevant span.

If you want to select a new answer, please click \textit{reset answer}, if you are ready to see the model answer, please click \textit{show answer}. Note that your answer will lock at that time.

\paragraph{Raw Results.}
In our evaluation, we use the first 15 points as training, therefore, we discard them from the main evaluation but show them in this section.  Overall, we see that training, for the most part has a positive effect, or not so much of an effect. These scores can be seen in Table \ref{tab:raw}.

\begin{table}[h]
\footnotesize
\begin{center}
\begin{tabular}{l| r |rrr}
\toprule

&{\sc Model}&\multicolumn{3}{c}{\sc Human}\\
\midrule 

& & \multicolumn{3}{c}{\sc \textbf{Baseline}} \\
\midrule 
 {\sc Low-orig}& 0.17
 & 0.14 & 0.52 &52.27 \\
 {\sc Low-Adv}& 0.15
 & 0.10 & 0.36&54.36\\
 \midrule
 {\sc High-orig} & 0.79
 & 0.53& 0.58 &37.12\\
 {\sc High-Adv}& 0.66%
  & 0.35& 0.48 & 47.64\\
 \midrule 
& & \multicolumn{3}{c}{\sc \textbf{Integrated (IG)}} \\
\midrule 
 {\sc Low-orig}& \cellcolor{gray}& $^*$0.34 & 0.35 &41.68 \\
{\sc Low-Adv}& \cellcolor{gray}& $^*$0.36 &0.28 &44.38 \\ 

\midrule 

{\sc High-orig} &\cellcolor{gray} &  $^*$0.71 & 0.76 &46.87  \\
{\sc High-Adv}& \cellcolor{gray}&  0.46& 0.47 & 42.99 \\
\midrule 
& & \multicolumn{3}{c}{\sc \textbf{Gradients}} \\
\midrule 
 {\sc Low-orig}& \cellcolor{gray}&  $^*$\textbf{0.64} & $^*$0.09 &$^*$32.16 \\
{\sc Low-Adv}& \cellcolor{gray}& $^*$\textbf{0.63} &0.23 &$^*$30.05 \\ 

\midrule 

{\sc High-orig} &\cellcolor{gray} & $^*$\textbf{0.82} & $^*$0.84 &44.65  \\
{\sc High-Adv}& \cellcolor{gray}&  $^*$\textbf{0.57}& $^*$0.62 &$^*$52.30 \\
\bottomrule
\end{tabular}
\caption{\label{tab:raw} Raw scores, before removing data points on training session }
\end{center}
\end{table}

\section{Experiment 2: Best Model Selection}
\label{e1-app}
Below we show the instructions given to the participants, and more details about the qualitative analysis of the feedback we obtained.

\paragraph{Instructions.}
\label{instructions1}
Question-answering (QA) systems are a particular form of artificial intelligence. We have trained two QA systems and have extracted the most important words the model uses to make its final decision. Based on these highlighted words, your task is to select the model that you think is more likely to perform best. Additionally, please write how the low-performing model fails and/or how it could be better (try to be detailed)

\begin{figure}[h]
\centering
\includegraphics[width=.5\textwidth]{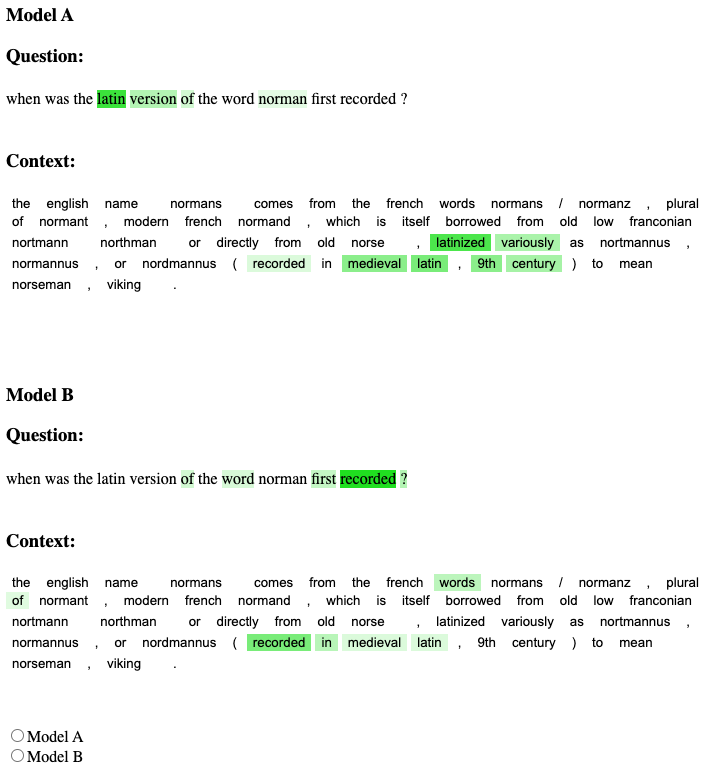}
\caption{\label{fig:e1}  Experiment 1 UI: {\sc Low}(bottom) vs {\sc High}(top) condition.}
\end{figure}

\paragraph{User Interface.}
An example instance, as shown to the participants, can be seen in \autoref{fig:e1}.

\begin{table*}[t]
\begin{center}
 \footnotesize
\begin{tabular}[h]{p{2.5cm} |p{10cm} }
\toprule

{\sc Qualitative Codes} & {\sc examples} \\
\hline

\multirow{2}{*}{Irrelevant (q)} & \textbf{1. } Model A only extracted some important words but also some punctuations in the question which is insufficient to derive to a good answer. Model B extracted a number of key important words that would lead to the correct answer.\\
& \textbf{2.} Option b chose quantitative statements, while option A seems confused about what it's looking for since it highlights all sorts of things in the question. \\
\hline
\multirow{3}{*}{main entity (q)} & \textbf{1.} The words "year" and "norman" in the question were not extracted by Model A. The Model will not be able get the correct answer without knowing what to look for. \\
& \textbf{2.} The question was asking about the year lavoisier's work was published but neither of the key words in this question were highlighted. Model A had no idea where to locate the answer without considering those key words.\\
 \hline
\multirow{2}{*}{main entity (a)} & \textbf{1.} The answer requires a year; it hasn't highlighted any years as part of the answer. \\
& \textbf{2.} Answer needed to be a name and option A chose nothing that could be a name.\\
 
 \hline
 
\multirow{3}{*}{Irrelevant (a)} & \textbf{1.} Model B has highlighted many extra words in the answer\\
& \textbf{2.} Both models selected the correct terms, but model A selected more irrelevant terms in the answer too, so it's less likely to choose the correct one from those numerous options. \\
& \textbf{3.} B highlighted the answer but also too much unneeded info. \\

  \hline
 
\multirow{2}{*}{Generic/correctness} & \textbf{1.}  Model A does not highlight the right answer\\
& \textbf{2.}  Model B is wrong and model A is correct\\

\bottomrule
\end{tabular}
\caption{\label{tab:coding-examples}  Examples of some of the feedback categorized into these classes }
\end{center}
\end{table*}

\paragraph{Qualitative analysis of feedback.}
In Table \ref{tab:coding-examples}, we include a few examples of the sentence that were categorized using the qualitative codes. Unsurprisingly, once participants found a strategy for giving feedback , they mostly stuck to it.

After categorizing all the feedback into each category, we visualize the distribution per condition. This can be found in Figure \ref{fig:distribution}. We find that for the {\sc High} vs {\sc Low} conditions, the distribution is very similar between gradients and integrated gradients. Many participants gave very generic feedback , for example by simply saying that "model A is better because it is correct, and model B is wrong". This was not surprising, as here the differences were supposed to be clear and it is likely most participants did not have to think too hard before making a decision.  However, the distribution is very different for the {\sc High} vs {\sc Medium} conditions. Here, for standard gradients, the feedback followed a similar pattern as in the previous condition, but about 30\% less examples received generic feedback than before. For integrated gradients, most examples received feedback regarding the irrelevant terms being highlighted, showing that even when the difference in performance between models is large (20 F1 points), this method makes the distinction difficult for the best model selection task.

\end{document}